\documentclass{article}
\usepackage{ijcai19}

\usepackage{times}
\usepackage{helvet}
\usepackage{courier}
\usepackage{amsmath}
\usepackage{amsthm}
\usepackage{amssymb}    
\usepackage{graphicx}
\usepackage{color}
\usepackage{url,cleveref}

\usepackage{algorithm}
\usepackage{algorithmic}
\usepackage{braket}
\usepackage{array}
\usepackage{tabu}
\usepackage{multirow,multicol,ctable}
\usepackage{subfigure}
\usepackage{epstopdf}
\title{Real-World Image Datasets for Federated Learning}

\author{
Jiahuan Luo\textsuperscript{\rm 1*}{\footnote{Authors performed the work while they were research interns at WeBank Co. Ltd., Shenzhen, China}}, 
Xueyang Wu\textsuperscript{\rm 2*}, 
Yun Luo\textsuperscript{\rm 2,3}, 
Yunfeng Huang\textsuperscript{\rm 4*},  
Yang Liu\textsuperscript{\rm 5}\footnote{Yang Liu is the corresponding author}, \\
Aubu Huang\textsuperscript{\rm 5}, 
Qiang Yang\textsuperscript{\rm 2,5}
\affiliations
\textsuperscript{\rm 1}South China University of Technology, China\\
\textsuperscript{\rm 2}Hong Kong University of Science and Technology, Hong Kong SAR, China\\
\textsuperscript{\rm 3}Extreme Vision Co. Ltd., Shenzhen, China\\
\textsuperscript{\rm 4}Shenzhen University, China\\
\textsuperscript{\rm 5}WeBank Co. Ltd., Shenzhen, China\\
\emails
seluojiahuan@mail.scut.edu.cn, \{xwuba, yluoav\}@cse.ust.hk, huangyunfeng2017@email.szu.edu.cn\\ \{yangliu, stevenhuang\}@webank.com, qyang@cse.ust.hk
}

\begin{document}

\maketitle

\begin{abstract}
Federated learning is a new machine learning paradigm which allows data parties to build machine learning models collaboratively while keeping their data secure and private. While research efforts on federated learning have been growing tremendously in the past two years, most existing works still depend on pre-existing public datasets and artificial partitions to simulate data federations due to the lack of high-quality labeled data generated from real-world edge applications. Consequently, advances on benchmark and model evaluations for federated learning have been lagging behind. In this paper, we introduce a real-world image dataset. The dataset contains more than 900 images generated from 26 street cameras and 7 object categories annotated with detailed bounding box. The data distribution is non-IID and unbalanced, reflecting the characteristic real-world federated learning scenarios. Based on this dataset, we implemented two mainstream object detection algorithms (YOLO and Faster R-CNN) and provided an extensive benchmark on model performance, efficiency, and communication in a federated learning setting. Both the dataset and algorithms are made publicly available\footnote{Dataset and code are available at \url{https://dataset.fedai.org} and \url{https://github.com/FederatedAI/research}}.
\end{abstract}

\section{Introduction}
Object detection is at the core of many real-world artificial intelligence (AI) applications, such as face detection \cite{Sung:1998:ELV:275341.275345}, pedestrian detection \cite{Dollar:2012:PDE:2197081.2197275}, safety controls, and video analysis. With the rapid development of deep learning, object detection algorithms have been greatly improved in the past few decades \cite{Ren:2017:FRT:3101720.3101780,DBLP:journals/corr/abs-1804-02767,DBLP:journals/corr/abs-1807-05511}. A traditional object detection approach requires collecting and centralizing a large amount of annotated image data. Image annotation is very expensive despite crowd-sourcing innovations, especially in areas where professional expertise is required, such as disease diagnose. In addition, centralizing these data requires uploading bulk data to database which incurs tremendous communication overhead. For autonomous cars, for example, it is estimated that the total data generated from sensors reach more than 40GB/s. Finally, centralizing data may violate user privacy and data confidentiality and each data party has no control over how the data would be used after centralization. In the past few years, there has been a strengthening private data protection globally, with law enforcement including the General Data Protection Regulation (GDPR) \cite{Voigt:2017:EGD:3152676}, implemented by the European Union on May 25, 2018. 

\begin{figure}[t]
\includegraphics[width=3.38in]{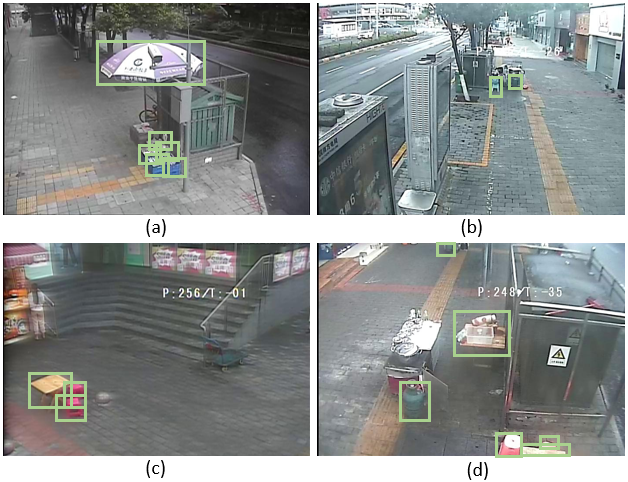}
\caption{Examples taken from Street Dataset. The green bounding boxes represent the target objects.}
\label{sample images}   
\end{figure}

To overcome the challenge of data privacy and security in the machine learning process, many privacy-preserving machine learning methods have been proposed in the past, such as secure multi-party computing (MPC) in the 1980s \cite{Yao:1982:PSC:1398511.1382751,Yao:1986:GES:1382439.1382944}. MPC allows multiple parties to collaboratively calculate a convention function safely without revealing their data to each other or a trusted third party. However, traditional MPC protocols require high communication overhead between parties, making it difficult to be applied to industry. Differential Privacy, proposed by Dwork in 2006 \cite{Dwork:2008:DPS:1791834.1791836}, protects user data by adding noise, but incurs a trade-off between model accuracy and the risk of data leakage.

Federated learning \cite{DBLP:journals/corr/McMahanMRA16}, an emerging machine learning paradigm, allows users to collaboratively train a machine learning model while protecting user data privacy. In essence, federated learning is a collaborative computing framework. The idea is to train through model aggregations rather than data aggregation and keep local data stay at the local device. Since most of data (images, videos, text...) are generated from edge devices, federated learning is an attractive approach to enable end-to-end computer vision tasks with image annotation and training tasks moved on the edge whereas only model parameters are sent to the central cloud for aggregation.  

\begin{table}[t]
\caption{The object category distribution over the Street Dataset}
\centering
\label{table-labels}
\begin{tabular}{ccc}
\specialrule{.1em}{.05em}{.05em} 
Object Category & Sample & Frequency \\ \specialrule{.1em}{.05em}{.05em} 
Basket & 162 & 211 \\ \hline
Carton & 164 & 275 \\ \hline
Chair & 457 & 619 \\ \hline
Electromobile & 324 & 662 \\ \hline
Gastank & 91 & 137 \\ \hline
Sunshade & 314 & 513 \\ \hline
Table & 88 & 127 \\ \specialrule{.1em}{.05em}{.05em} 
\end{tabular}
\end{table}

 \begin{figure}[t] 
	\centering
	\includegraphics[width=3.2in]{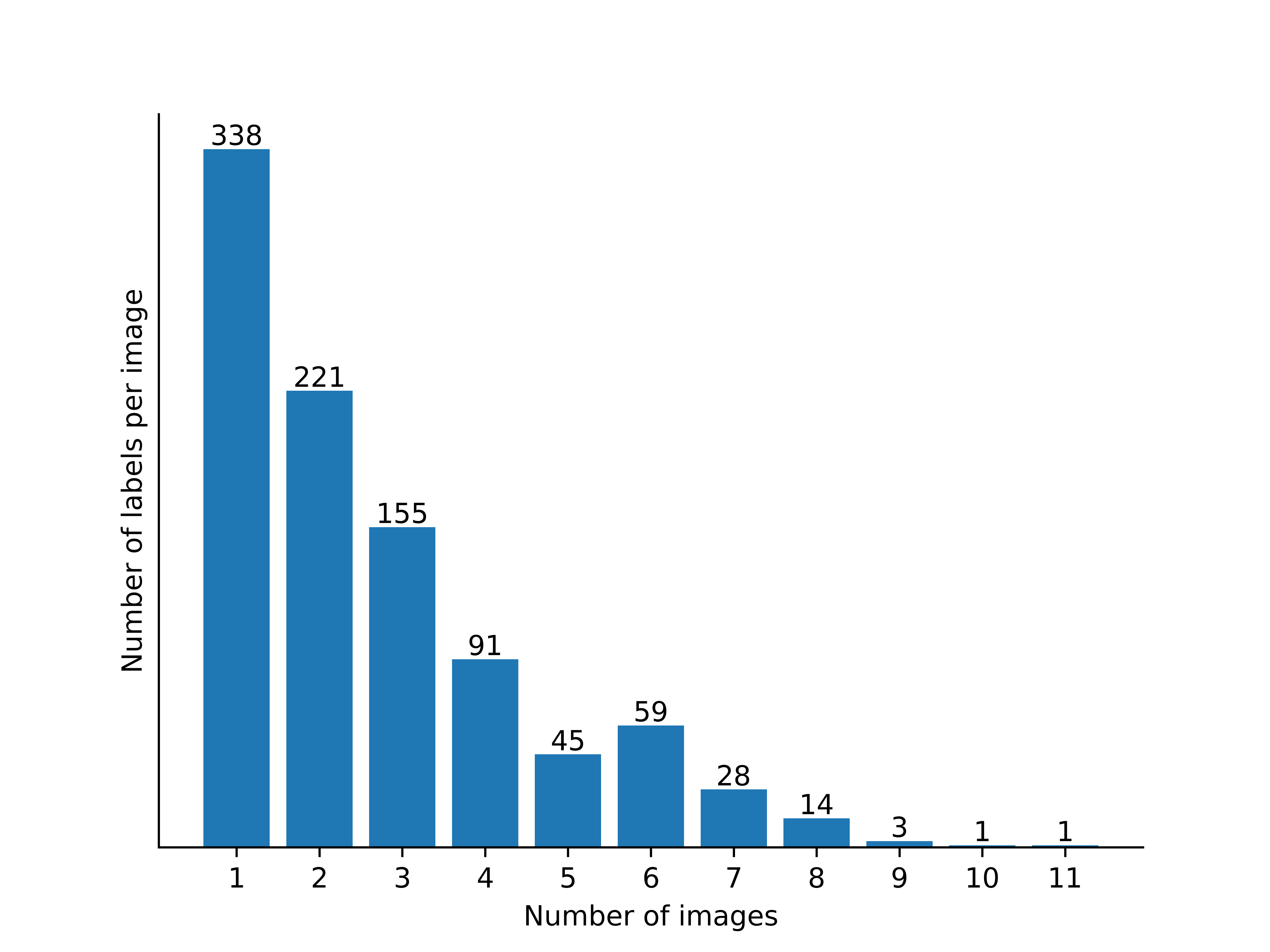}
	\caption{The distribution of labels per image.}
	\label{fig-distribution_labels} 
\end{figure}

 \begin{figure}[t] 
	\centering
	\includegraphics[width=3.5in, trim={5 5 5 5},clip]{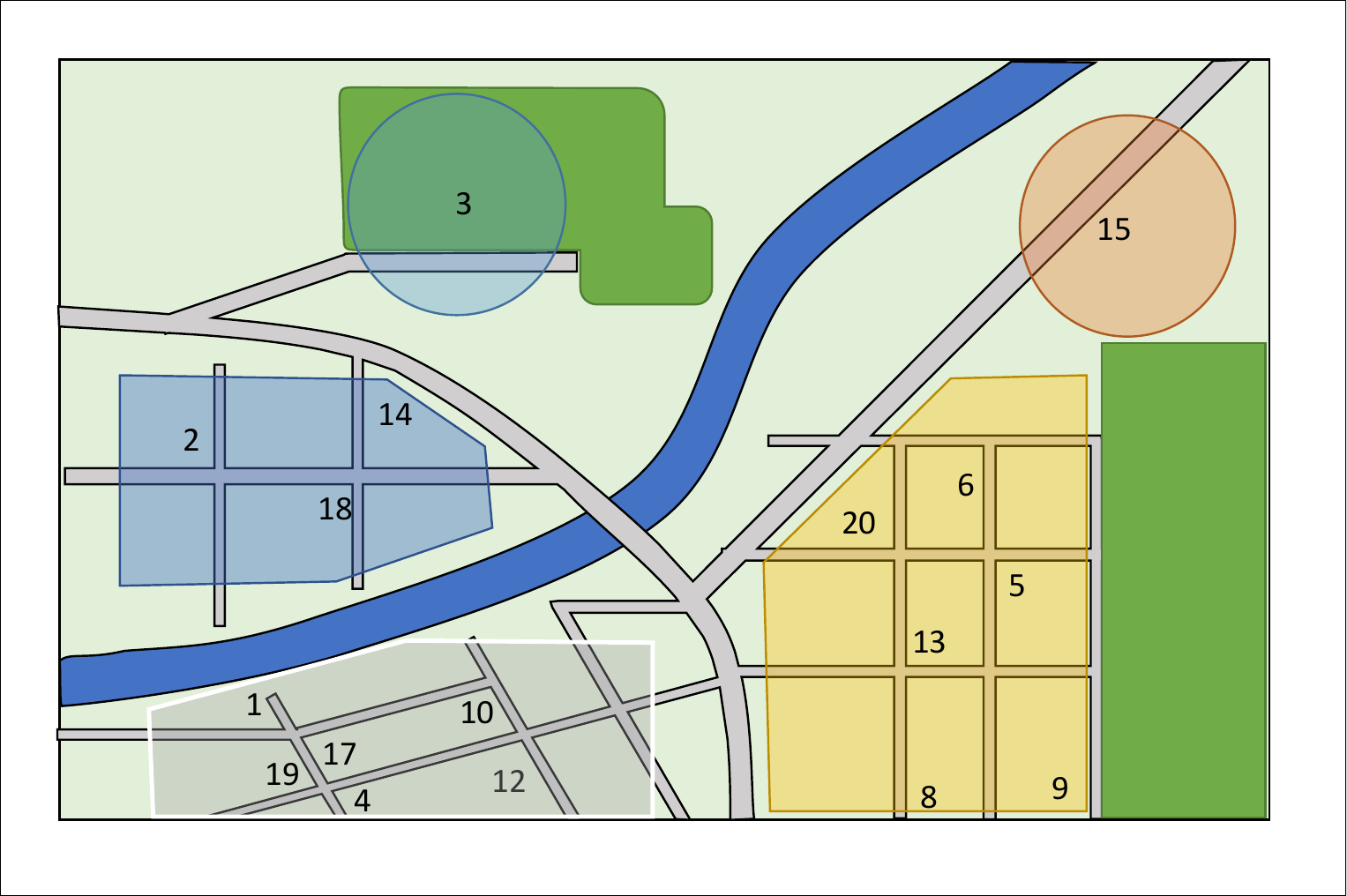}
	\caption{The schematic view of cameras' geometric information.}
	\label{location} 
\end{figure}

Despite the rapid growth of research interest on federated learning, current research work still rely on the pre-existing public datasets which are designed for centralized machine learning. There is a lack of real-world edge datasets representing the truly non-IID and unbalanced data distributions in federated learning. Consequentially, there are still significant lags in model evaluations and benchmark for federated learning. 
In this paper, we present a real-world image dataset generated from street cameras and then manually annotated and selected.  The dataset is a realistic representation of real-world image data distributions and therefore are non-IID and unbalanced. We carefully evaluate these images and analyze the detailed statistics of the object distributions. In addition, we implement two state-of-the-art object detection algorithms, YOLOv3 and Faster R-CNN, and integrate them with federated learning technique. We evaluate the model performance comprehensively and provide baselines and comparisons using this federated dataset. We make this dataset and evaluation source code public that will be available.

\section{Related Work}

As an emerging technology, federated learning introduces new challenges in system design, communication efficiency and model evaluation. Google presented an overview of federated learning system design in 2019 \cite{47976}, highlighting some of the design challenges for large-scale problems. The complexity of federated learning systems poses challenges for both implementation and experimental evaluation. As a result, many open-source projects were proposed to lower this barrier to entry. TensorFlow Federated (TFF), led by Google is an open-source framework on top of TensorFlow for flexibly expressing arbitrary computation on decentralized data. Federated AI Technology Enabler (FATE) led by WeBank is an open-source technical framework that enables distributed and scalable secure computation protocols based on homomorphic encryption and multi-party computation. Moreover, OpenMined proposes a python Library for secure, private deep learning \cite{ryffel2018generic}. It provides a general frameowrk to incoorporate federated learning with deep learning models implemented with PyTorch \cite{paszke2017pytorch}. These federated learning framework allows researchers to explore federated learning algorithms and applications in a more convenient and efficient way. LEAF \cite{DBLP:journals/corr/abs-1812-01097} is a benchmark framework that contains preprocessed datasets, each with a ``natural'' partitioning that aims to reflect the type of non-identically distributed data partitions encountered in practical federated environments. 

So far, federated learning has been implemented with many state-of-the-art algorithms such as long short-term memory (LSTM) networks \cite{mcmahan2017learning8},  support vector machines (SVMs) \cite{rubinstein2009learning}, gradient boosting trees \cite{cheng2019secureboost46}, etc.. Federated learning algorithms are also applied to industrial scenarios including next-word-prediction \cite{hard2018federated28}, speech keyword spotting \cite{leroy2018federated15}, and images classification \cite{liu2018secure45}. \cite{chen2019federated78} further improves next-word-prediction by addressing the out-of-vocabulary problem in language model personalization.  To the best of our knowledge, this is the first time that object detection algorithms are implemented with federated learning providing a reliable benchmark framework and end-to-end solutions for real-world vision tasks. 

As one of the fundamental problems of computer vision, object detection forms the basics of many other computer vision tasks, such as instance segmentation \cite{hariharan2014simultaneous}, image caption \cite{karpathy2015deep}, object tracking \cite{kang2017t}, etc. Object detection is able to provide valuable information for semantic understanding of images and videos, and is related to many applications, including face recognition, pedestrian detection and autonomous driving \cite{chen2015deepdriving}. In recent years, the rapid development of deep learning techniques has brought new blood into object detection, leading to remarkable breakthroughs and making it widely used in many real-world applications.

\section{Street Dataset Description}

\subsection{Source Description}
We randomly capture these images of different scenes at different time from 26 street monitoring videos with $704\times576$ pixels. Similar and night scene images are removed from them. The remaining 956 images are legible and have obvious distinction in content. Eventually, we select a total of 2,544 items from these images with 7 object categories. Each image has at least one labeled object, and may have multiple labels of this same category in one image. 
The object labels are basket, carton, chair, electromobile, gastank, sunshade and table. The distribution of these object labels is shown in Table~\ref{table-labels}, which demonstrates that the class distribution is unbalanced. In addition, we also calculate the object frequency per image, shown in Figure~\ref{fig-distribution_labels}.

\begin{table}[!htp]
\caption{Detailed distribution of the object labels in the training set of Street-5 Dataset and Street-20 Dataset}\label{tab:street5}
\centering
\small
\begin{tabular}{crrr|rrr}
\specialrule{.1em}{.05em}{.05em} 
\multirow{2}{*}{\begin{tabular}[c]{@{}c@{}}Street-5 \\ Dataset\end{tabular}} & \multicolumn{3}{c}{Images/Client} & \multicolumn{3}{c}{Frequency/Client} \\ \cline{2-7} 
 & \multicolumn{1}{c}{total} & \multicolumn{1}{l}{mean} & \multicolumn{1}{c}{stdev} & \multicolumn{1}{c}{total} & \multicolumn{1}{c}{mean} & \multicolumn{1}{c}{stdev} \\ \specialrule{.1em}{.05em}{.05em} 
Basket & 127 & 25.40 & 15.08 & 165 & 33.00 & 22.20 \\ \hline
Carton & 133 & 26.60 & 27.94 & 215 & 43.00 & 55.13 \\ \hline
Chair & 369 & 73.80 & 59.36 & 494 & 98.80 & 87.60 \\ \hline
\begin{tabular}[c]{@{}c@{}}Electro\\ -mobile\end{tabular} & 257 & 51.40 & 45.23 & 510 & 102.00 & 105.27 \\ \hline
Gastank & 71 & 14.20 & 26.42 & 106 & 21.20 & 36.19 \\ \hline
Sunshade & 255 & 51.00 & 31.89 & 413 & 82.60 & 56.02 \\ \hline
Table & 73 & 14.60 & 29.20 & 102 & 20.40 & 40.80 \\ 
 \specialrule{.1em}{.05em}{.05em} 
 \\
 \specialrule{.1em}{.05em}{.05em} 
\multirow{2}{*}{\begin{tabular}[c]{@{}c@{}}Street-20 \\ Dataset\end{tabular}} & \multicolumn{3}{c}{Images/Client} & \multicolumn{3}{c}{Frequency/Client} \\ \cline{2-7} 
 & \multicolumn{1}{c}{total} & \multicolumn{1}{l}{mean} & \multicolumn{1}{c}{stdev} & \multicolumn{1}{c}{total} & \multicolumn{1}{c}{mean} & \multicolumn{1}{c}{stdev} \\ \specialrule{.1em}{.05em}{.05em} 
Basket & 127 & 6.35 & 11.06 & 165 & 8.25 & 14.93 \\ \hline
Carton & 133 & 6.65 & 12.03 & 215 & 10.75 & 26.23 \\ \hline
Chair & 369 & 18.45 & 17.70 & 494 & 24.70 & 28.54 \\ \hline
\begin{tabular}[c]{@{}c@{}}Electro\\ -mobile\end{tabular} & 257 & 12.85 & 14.81 & 510 & 25.50 & 37.55 \\ \hline
Gastank & 71 & 3.55 & 11.14 & 106 & 5.30 & 16.63 \\ \hline
Sunshade & 255 & 12.75 & 19.14 & 413 & 20.65 & 37.87 \\ \hline
Table & 73 & 3.65 & 13.06 & 102 & 5.10 & 19.26 \\ \specialrule{.1em}{.05em}{.05em} 
\end{tabular}
\end{table}

 \begin{figure}[ht] 
	\centering
	\includegraphics[width=3.4in]{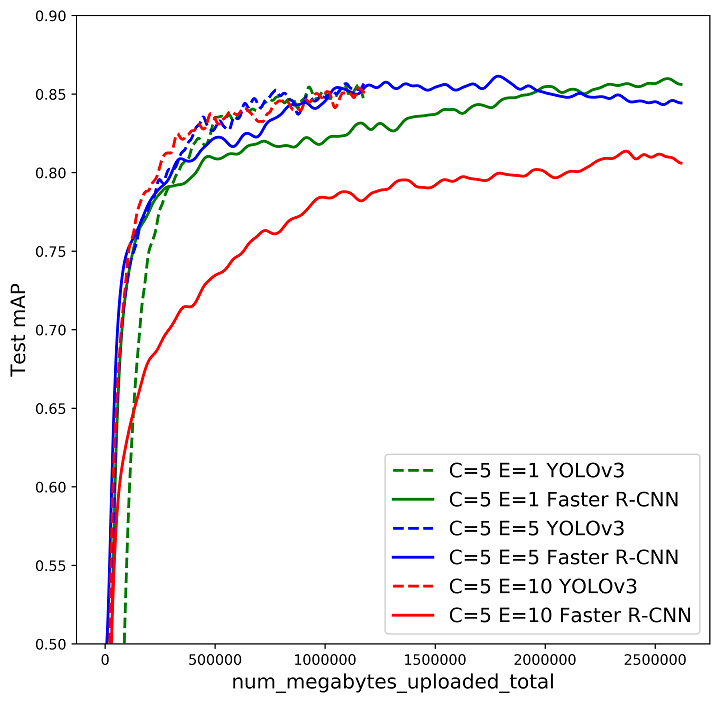}
	\caption{Test set mAP vs number of megabytes uploaded for different models.}
	\label{uploaded_map} 
\end{figure}

Visual objects are annotated with bounding boxes. We use bounding box of description ($x_{min}, y_{min}, x_{max}, y_{max}$), where ($x_{min}, y_{min}$) is the top-left coordinates of the bounding box and ($x_{max}, y_{max}$) the bottom-right.

\subsection{Data Division}
We split the Street Dataset according to the geographic information\footnote{Notably, the geographic information is hashed when the dataset is published.} of the camera. We naturally have 26 street monitoring videos, but some of the cameras have very few images. We split the whole dataset with 956 images into around 80\% (765 images) for training and 20\% (191 images) for testing. The testing set consists of images from 6 cameras with very few images, and random sample from other 20 cameras. By splitting the dataset this way, our testing set is able to jointly evaluate the predictability (with images from existed cameras) and generalization capability (with images from unseen cameras). 

We further divide the remaining training set into several clients according to its real world attribute, i.e., geographic information. Figure \ref{location} reflects the schematic location information of each device, denoted as device ID. We consider two scenarios: 1) Each camera device is a client, named as Street-20; 2) Nearby cameras are grouped into a single client to form in total 5 clients, named as Street-5. And we publish two datasets based on this method of division. Street-5 mimics the scenario where several edge devices are communicated and controled by a central node. Both datasets share the same testing set for evaluation. 
\begin{table*}[t]
\caption{Detailed distribution of the object frequency on each client in the training set of Street-5 Dataset and Street-20 Dataset}\label{tab:client2class}
\centering
\small
\begin{tabular}{crrrrrrr}
\specialrule{.1em}{.05em}{.05em} 
Client ID & Basket & Carton & Chair & Electromobile & Gastank & Sunshade & Table \\
\specialrule{.1em}{.05em}{.05em} 
1 & 23 & 145 & 148 & 209 & 12 & 35 & \textbf{0}\\\hline
2 & 64 & 57 & 249 & 249 & 93 & 102 & 102\\\hline
3 & \textbf{0} & \textbf{0} & 38 & 12 & \textbf{0} & 1 & \textbf{0}\\\hline
4 & 50 & 13 & 36 & 40 & 1 & 151 & \textbf{0}\\\hline
5 & 28 & \textbf{0} & 23 & \textbf{0} & \textbf{0} & 124 & \textbf{0}\\
\hline\hline
1 & 33 & 35 & 124 & 44 & 75 & 76 & 88\\\hline
2 & 50 & 13 & 21 & 40 & \textbf{0} & 119 & \textbf{0}\\\hline
3 & 28 & \textbf{0} & 23 & \textbf{0} & \textbf{0} & 124 & \textbf{0}\\\hline
4 & \textbf{0} & \textbf{0} & 50 & \textbf{0} & 18 & \textbf{0} & \textbf{0}\\\hline
5 & 22 & 2 & 51 & 144 & \textbf{0} & \textbf{0} & \textbf{0}\\\hline
6 & 1 & \textbf{0} & 2 & 14 & \textbf{0} & 1 & \textbf{0}\\\hline
7 & \textbf{0} & 116 & 21 & \textbf{0} & \textbf{0} & 18 & \textbf{0}\\\hline
8 & \textbf{0} & \textbf{0} & \textbf{0} & 11 & \textbf{0} & \textbf{0} & \textbf{0}\\\hline
9 & \textbf{0} & \textbf{0} & \textbf{0} & 40 & \textbf{0} & 2 & \textbf{0}\\\hline
10 & \textbf{0} & \textbf{0} & \textbf{0} & 92 & \textbf{0} & \textbf{0} & \textbf{0}\\\hline
11 & \textbf{0} & \textbf{0} & \textbf{0} & 77 & \textbf{0} & \textbf{0} & \textbf{0}\\\hline
12 & 31 & \textbf{0} & \textbf{0} & 11 & \textbf{0} & \textbf{0} & \textbf{0}\\\hline
13 & \textbf{0} & 27 & 44 & \textbf{0} & \textbf{0} & \textbf{0} & \textbf{0}\\\hline
14 & \textbf{0} & \textbf{0} & 13 & \textbf{0} & 1 & \textbf{0} & \textbf{0}\\\hline
15 & \textbf{0} & \textbf{0} & 38 & 12 & \textbf{0} & 1 & \textbf{0}\\\hline
16 & \textbf{0} & 22 & 22 & 18 & \textbf{0} & 11 & \textbf{0}\\\hline
17 & \textbf{0} & \textbf{0} & 38 & 1 & \textbf{0} & \textbf{0} & 14\\\hline
18 & \textbf{0} & \textbf{0} & 2 & \textbf{0} & \textbf{0} & 32 & \textbf{0}\\\hline
19 & \textbf{0} & \textbf{0} & 15 & 6 & \textbf{0} & 15 & \textbf{0}\\\hline
20 & \textbf{0} & \textbf{0} & 30 & \textbf{0} & 12 & 14 & \textbf{0}\\\specialrule{.1em}{.05em}{.05em} 
\end{tabular}
\end{table*}

\subsubsection{Street-5}
We cluster the remaining cameras into 5 groups, each of which represents as a client in the federated learning, according to the following considerations:
\begin{enumerate}
    \item The cameras should be clustered according to their geographic information, i.e., a client possesses the images from nearby cameras or cameras from the same region. 
    \item The number of samples in each client should have large divergence. 
\end{enumerate}

This split aims to simulate the federated learning running on different monitoring camera owners, such as different security companies. In this case, each company usually has more than one nearby cameras, and they can contribute to the federated learning with more images as a whole. 

\subsubsection{Street-20}
This dataset division is based on the minimal unit of our raw dataset, which aims to simulate the case where federated learning algorithms run on each device. In this case, the data are kept and processed within each client with the minimal risk to reveal the raw data. 

Since our data division is based on real-world distribution of cameras, our datasets suffer from non-IID data distribution problem. Table \ref{tab:street5} shows the detailed distributions of the distribution of annotated boxes among different clients, from which we can derive the unbalanced distribution of boxes, which may lead to learning bias of each client. From Table \ref{tab:client2class} we can learn the number of classed in each client more intuitively. Therefore, our published datasets can serve as good benchmarks for researchers to examine their federated learning algorithm's ability to address the non-IID distribution problem in real-world applications. 

\section{Experiments}
We evaluate two object detection methods on the proposed datasets as our benchmark algorithms, including Faster R-CNN\footnote{based on the implementation \url{https://github.com/chenyuntc/simple-faster-rcnn-pytorch}} \cite{ren2015faster}  and YOLOv3\footnote{based on the implementation \url{https://github.com/eriklindernoren/PyTorch-YOLOv3}} \cite{redmon2018yolov3} for their excellent performance. Note that, the backbone networks of Faster R-CNN is VGG16 \cite{simonyan2014very} and Darknet-53 for YOLOv3. 

\subsection{Baseline Implementation}
The code of our benchmark will be released, where two object detection models are implemented using PyTorch \cite{paszke2017pytorch}, on a GPU server with CPU of Intel Xeon Gold 61xx and 8 GPUs of Tesla V100.

Training Faster R-CNN was via SGD with a fixed learning rate of 1e-4 and a momentum of 0.9, while training YOLOv3 was via Adam with an initialization learning rate of 1e-3. Notably, we use pretrained VGG16 model for Faster R-CNN for faster convergence. In terms of model size, the YOLOv3 model has 61,556,044 parameters, and the Faster R-CNN model has 137,078,239 parameters with backbone network of VGG16.

We adapt the original \texttt{FederatedAveraging} (\texttt{FedAvg}) algorithm \cite{DBLP:journals/corr/McMahanMRA16} to framework, shown in Algorithm \ref{algo_train}. As our purpose is to examine the effect of different data division and federated learning settings, we modified \texttt{FedAvg} algorithm to a \textit{pseudo} \texttt{FedAvg} algorithm, by replacing the server-client communication framework such as SocketIO with saving and restoring checkpoints on hard-devices, which simplifies the processing of model aggregation. However, our implementation can also be easily migrated to \texttt{FedAvg}. 

There are three key parameters of the \texttt{FedAvg} algorithm: $C$, the number of clients that participate in training on each round; $E$ , the number of times that each client performs over its local dataset on each round; and $B$, the minibatch size used for client updates. All these three parameters control the computation. We mainly set $B$ = 1 for the experiments when running Faster R-CNN.

As for \texttt{FedAvg} algorithm, it can select a $R$-fraction of clients on each round, and compute the gradient of the loss over all the data held by these clients. Thus, in this algorithm $R$ controls the global batch size, with $R$ = 1 corresponding to full-batch gradient descent. In order to produce a better result, we fix $R$ = 1, which means the server waits for the updates of all the clients participating in training.

\begin{algorithm}[!htp]
\small
\caption{Pseudo FedAvg}
\begin{algorithmic}\label{algo_train}
\REQUIRE $N$ client parties $\{c_k\}_{k=1..N}$, total rounds $T$, and Server side $\mathcal{S}$;
\ENSURE Aggregated Model $w$
\STATE ~~\ 
 $\mathcal{S}$ initializes federated model parameters, and saves as checkpoint. Client parties $\{c_k\}_{k=1..N}$ load the checkpoints. 
\FOR{$t=1,...,T$}
    \FOR{$k=1,...,N$}
        \STATE {$w_k = w^{(t)}$}
        \STATE each client $\{c_k\}$ do local training:
        \FOR{$i=0,1,...,M_k$}
            \STATE \textit{\small{($M_k$ is the number of data batches $b$ in the client $c_k$)}}
            \STATE client $\{c_k\}$  computes gradients $\nabla \ell(w_k, b_i)$
            \STATE update with $w_k = w_k - \eta\nabla \ell(w_k, b_i)$
        \ENDFOR
        \STATE save $w_k$ results to checkpoints.
    \ENDFOR
    \STATE $\mathcal{S}$ loads checkpoints and get averaged model with $w^{(t)} = \frac{1}{N} \sum_{k=1}^{N}{w_k}$
\ENDFOR
\RETURN $w^{(T)}$
\end{algorithmic}
\end{algorithm}

\begin{table}[t]
\caption{Number of communication rounds to reach a target mAP (75\%)}
\label{tab:given map}
\centering
\begin{tabular}{ccccc}
\specialrule{.1em}{.05em}{.05em}  
Model & $C$ & $E$ & Rounds & Amount (MB) \\ \specialrule{.1em}{.05em}{.05em} 
\multirow{6}{*}{\begin{tabular}[c]{@{}c@{}}\textbf{YOLOv3} \\ (w/o pretrained)\end{tabular}} 
    & 5 & 1 & 158 & 186,440 \\ \cline{2-5}
    & 5 & 5 & 48 & 56,640\\ \cline{2-5}
    & 5 & 10 & 90 & 106,200\\ \cline{2-5}
    & 20 & 1 & 83 & 391,760 \\ \cline{2-5}
    & 20 & 5 & 448 & 2,114,560 \\ \cline{2-5}
    & 20 & 10 & 346 & 1,633,120 \\ \cline{2-5}
    \specialrule{.1em}{.05em}{.05em} 
\multirow{6}{*}{\textbf{Faster R-CNN}} 
    & 5 & 1 & 45 & 117,675\\ \cline{2-5}
    & 5 & 5 & 30 & 78,450\\ \cline{2-5}
    & 5 & 10 & 189 & 494,235 \\ \cline{2-5}
    & 20 & 1 & 161 & 1,684,060 \\ \cline{2-5}
    & 20 & 5 & 119 & 1,244,740\\ \cline{2-5}
    & 20 & 10 & 90 & 941,400\\ \specialrule{.1em}{.05em}{.05em}
\end{tabular}
\end{table}

\subsection{Evaluation Metrics}
In this section, we summarize some common metrics for object detection in our federated learning framework.

\textbf{Intersection Over Union } Intersection Over Union (IOU) is used to evaluate the overlap between two bounding boxes. It requires a ground truth bounding box $B_{gt}$  and a predicted bounding box $B_p$. By applying the IOU we can tell if a detection is valid (True Positive) or not (False Positive). IOU is given by the overlapping area between the predicted bounding box and the ground truth bounding box divided by the area of union between them:  
\begin{gather*}
IOU=\frac{area(B_{gt} \cap B_{p})}{area(B_{gt} \cup B_{p})}
\end{gather*}

\textbf{mean Average Precision } We choose the standard PASCAL VOC 2010 mean Average Precision (mAP) for evaluation, in which mean is taken over per-class APs:
\begin{gather*}
    mAP=\frac{\sum_{i=1}^{k}AP_i}{k}
\end{gather*}
where Average Precision (AP) is calculated for each class respectively, $k$ is number of classes. 



\begin{figure*}[!ht]
    \centering
    \subfigure[YOLOv3]{
        \includegraphics[width=0.315\textwidth]{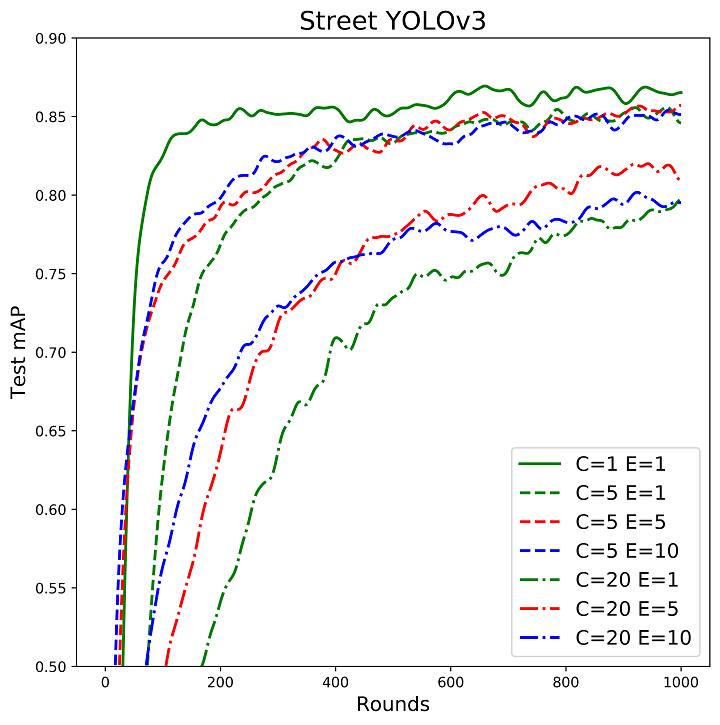}
        \label{fig:street_yolo_map}
    }
    \subfigure[Pretrained YOLOv3]{
        \includegraphics[width=0.315\textwidth]{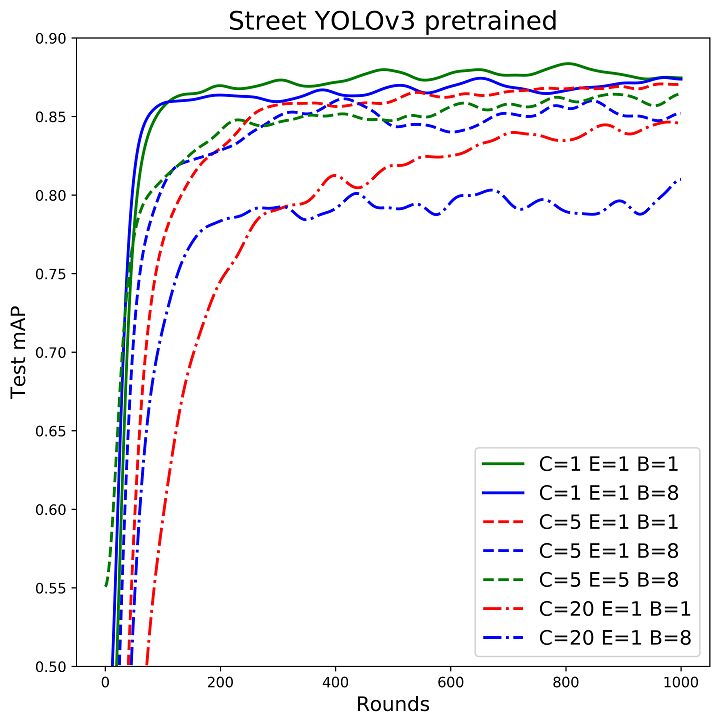}
        \label{fig:street_yolo_pretrain_map}
    }
    \subfigure[Pretrained Faster R-CNN]{
        \includegraphics[width=0.315\textwidth]{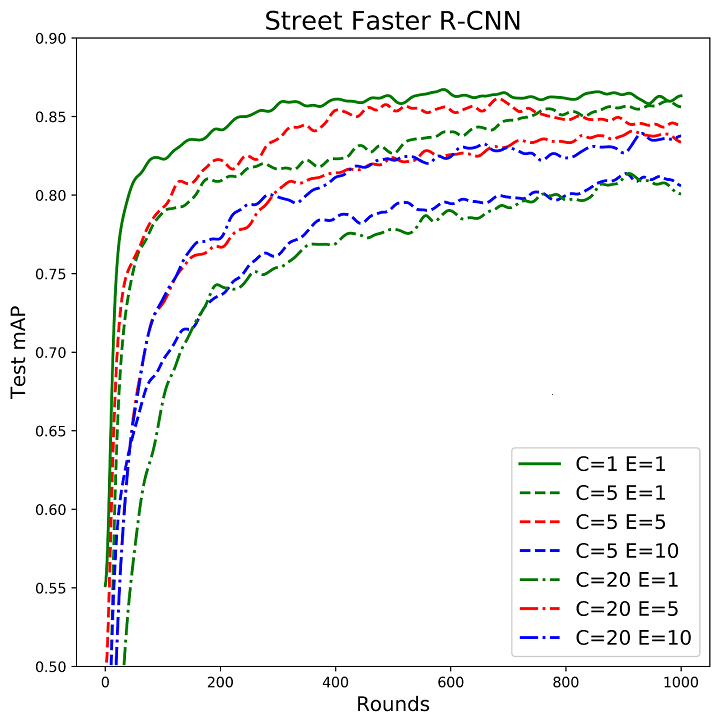}
        \label{fig:street_faster_map}
    }
    \caption{Test set mAP vs. number of communication rounds using different models}
    \label{fig:street-map}
\end{figure*}

\begin{figure*}[!ht]
    \centering
    \subfigure[YOLOv3]{
        \includegraphics[width=0.315\textwidth]{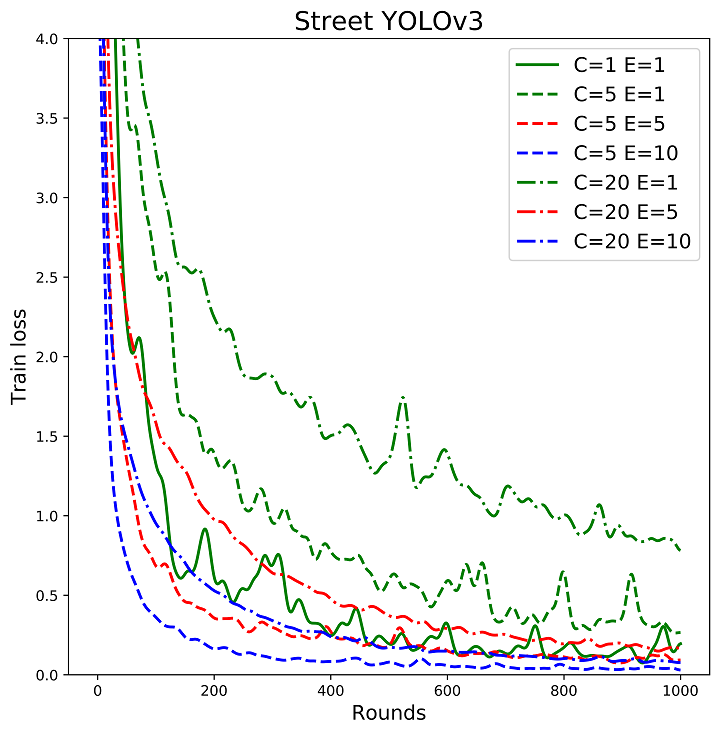}
        \label{fig:yolo_loss}
    }
    \subfigure[Pretrained YOLOv3]{
        \includegraphics[width=0.315\textwidth]{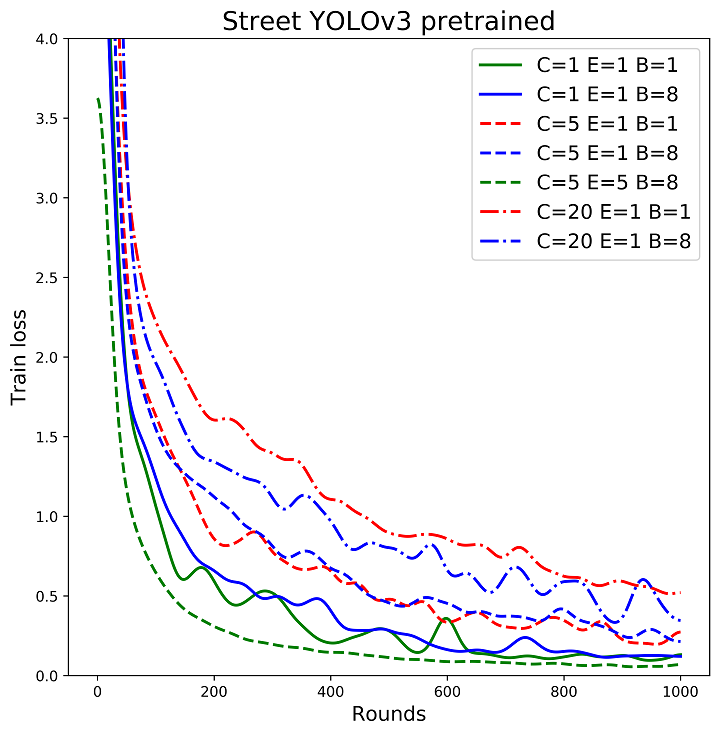}
        \label{fig:pretrain_yolo_loss}

    }
    \subfigure[Pretrained Faster R-CNN]{
        \includegraphics[width=0.315\textwidth]{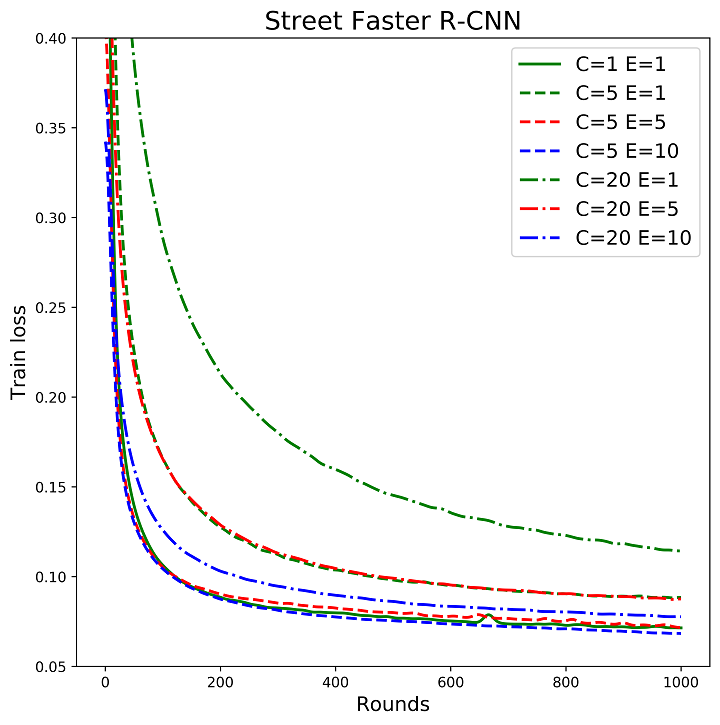}
        \label{fig:faster_loss}
    }
    \caption{Training loss vs. number of communication rounds using different models}
    \label{fig:train_loss}
\end{figure*}

\subsection{Results}
In this section, we report the baseline results. In order to thoroughly investigate the hyper-parameters of the \texttt{FedAvg} algorithm and evaluate the effect of different parameters, we conducted experiments in different parameter configurations. Note that, we use a threshold of 0.5 for IOU in our experiments when computing mAP.

Figure \ref{fig:street_yolo_map} shows the test-set mAP for YOLOv3 without pretrained model. We fix ${C}$ = 5, which means the data is split into five parts according to the geographic location, add more computation per client by increasing $E$, which controls the number of training passes. We see that increasing local epoch $E$ is effective. Figure \ref{fig:street_yolo_map} demonstrates that adding more local SGD updates per round and then averaging the resulting models can reach a higher mAP at the beginning of the training procedure. We report the number of communication rounds necessary to achieve a target test-set mAP. To achieve this, we evaluate the averaged model on each round to monitor the performance. Table \ref{tab:given map} quantifies this speedups. 

Expectantly, the \texttt{FedAvg} algorithm should have the same performance as centralized training. When it comes to non-IID datasets, it is difficult for \texttt{FedAvg} to reach the same score as that of centralized. Using the more non-IID dataset, setting $C$ = 20, shows a lower performance compared to the $C$ = 5 one. When we fix $C$ = 5, we get a comparable performance compared to centralized training. Though we stopped training at a number of communication rounds of 1000, it seems that the algorithm has not converged and they can get higher mAP if the training procedure continues. For both $C$ = 5 and $C$ = 20, larger $E$ usually converges faster. But as the training procedure goes on, when it comes to more non-IID cases, different $E$ leads to different performance and not the largest $E$ gets the best result. This result suggests that for non-IID datasets, especially in the later stages of convergence, it may be useful to decay the amount of local computation per round if we start at a large $E$ to lower communication costs.

The initial success of deep learning in computer vision can be largely attributed to transfer learning. ImageNet pretraining was crucial to obtain improvements over state-of-the-art results. Due to the importance of pretraining, we conduct additional experiments with pretrained models. Figure \ref{fig:street_yolo_pretrain_map} demonstrates that initialing with pretrained model weights produces a significant and stable improvement, especially for the Street-20 dataset, which has a small amount of pictures on each client. This shows that pretraining on large datasets is crucial for fine-tuning on small detection datasets. Furthermore, Figure \ref{fig:street_yolo_pretrain_map} shows the impact of batch size of each client. It is not very effective when we increase the batch size for each client. We conjecture this is largely due to the amount of pictures on each client is small. Especially on the Street-20 dataset, larger batch size even leads to lower performance, since each client contains only dozens of pictures. 

In addition to one-stage approach towards object detection, we contain Faster R-CNN as our benchmark, which is a popular two-stage approach. Figure \ref{fig:street_faster_map} reports the performance for Faster R-CNN with backbone network of pretrained VGG-16. For Faster R-CNN the $C$ = 5, $E$ = 1, \texttt{FedAvg} model eventually reaches almost the same performance as the centralized training, while the $C$ = 5, $E$ = 5, \texttt{FedAvg} model reaches a considerable mAP after 400 rounds. Training with pretrained model shows faster convergence. With $C$ = 5, small local epoch got better performance.

We also compare the training loss of different models. As shown in Figure \ref{fig:train_loss}, \texttt{FedAvg} is effective at optimizing the training loss as well as the generalization performance. Note the $y$-axes of different models are on different scales and loss is the average of all the clients. From Figure \ref{fig:yolo_loss}, we can see that in training, large local epoch $E$ always produces small loss and smooth training loss curve. We observed similar behavior for all three models. This is reasonable, because for large numbers of local epochs client would over-optimize on local dataset. One might conjecture large numbers of local epochs would bring about over-fitting. But they eventually reach a fairly similar mAP. Interestingly, for all three models, training with \texttt{FedAvg} converges to a high level of mAP. This trend continues even if the lines are extended beyond the plotted ranges. For example, for the YOLOv3 the $C$ = 5, $E$ = 1, \texttt{FedAvg} model reaches 88.86\% mAP after 1400 rounds, which is the best performance of centralized training.

We are also concerned with the communication costs when using different models. We choose the Faster R-CNN  as our cumbersome model and YOLOv3 as our lightweight model. The size of the parameters of Faster R-CNN is more than twice that of YOLOv3. Note that the backbone network of Faster R-CNN is VGG16. Figure \ref{uploaded_map} and Table \ref{tab:given map} demonstrate the communication rounds and costs to reach a target mAP of different models.

The unbalanced and non-IID distribution of the datasets are representative of the kind of data distribution for real-world applications. Encouragingly, it is impressive that naively average the parameters of models trained on clients respectively provides considerable performance. We conjecture that tasks like object detection and speech recognition, which usually require cumbersome model, are suitable and show significant result on Federated Learning.

\section{Conclusions and Future Work}
In this paper we release a real-world image dataset to evaluate federated object detection algorithms, with reproducible benchmark on Faster R-CNN and YOLOv3. Our released dataset contains common object categories on the street, which are naturally collected and divided according to the geographical information of the cameras. The dataset also captures the realistic non-IID distribution problem in federated learning, so it can serve as a reliable benchmark for further federated learning research on how to alleviate the non-IID problem. In the future, we will keep augmenting the dataset as well as presenting more benchmarks on these datasets.

\small

\bibliographystyle{named}
\bibliography{ref}

\end{document}